\documentclass[10pt, a4paper]{article}

\usepackage{lrec-coling2024} 

\usepackage{multirow}
\usepackage{multicol}
\usepackage{booktabs}
\usepackage{graphicx}

\title{Rewiring the Transformer with Depth-Wise LSTMs}

\name{Hongfei Xu$^{1,2}$, Yang Song$^1$, Qiuhui Liu$^3$, Josef van Genabith$^2$\thanks{Corresponding author: Josef van Genabith.}, Deyi Xiong$^4$} 

\address{$^1$Zhengzhou University, Henan, China\\
$^2$DFKI and Saarland University, Informatics Campus, Saarland, Germany\\
$^3$China Mobile Online Services, Henan, China\\
$^4$College of Intelligence and Computing, Tianjin University, Tianjin, China\\
\{hfxunlp, ysongnlp, liuqhano\}@foxmail.com, josef.van\_genabith@dfki.de, dyxiong@tju.edu.cn\\}

\abstract{
Stacking non-linear layers allows deep neural networks to model complicated functions, and including residual connections in Transformer layers is beneficial for convergence and performance. However, residual connections may make the model ``forget'' distant layers and fail to fuse information from previous layers effectively. Selectively managing the representation aggregation of Transformer layers may lead to better performance. In this paper, we present a Transformer with depth-wise LSTMs connecting cascading Transformer layers and sub-layers. We show that layer normalization and feed-forward computation within a Transformer layer can be absorbed into depth-wise LSTMs connecting pure Transformer attention layers. Our experiments with the 6-layer Transformer show significant BLEU improvements in both WMT 14 English-German / French tasks and the OPUS-100 many-to-many multilingual NMT task, and our deep Transformer experiments demonstrate the effectiveness of depth-wise LSTM on the convergence and performance of deep Transformers.
 \\ \newline  }

\begin{document}

\maketitleabstract

\section{Introduction}

The multi-layer structure together with non-linear activation functions allow neural networks to model complicated functions. Increasing the depth of models can increase their capacity and benefit their performance if optimization difficulties \cite{mhaskar2016learning,telgarsky2016benefits,eldan2016power,he2016deep,bapna2018training} can be properly addressed.

For machine translation, the performance of the Transformer translation model \cite{vaswani2017attention} benefits from including residual connections \cite{he2016deep} in stacked layers and sub-layers \cite{bapna2018training,wu2019depth,wei2020multiscale,zhang2019improving,xu2020lipschitz,li2020shallow,huang2020improving,xiong2020on,mehta2021delight,li2021learning,xu2021optimizing}. However, the residual connections within each layer only fuse information through simple, one-step operations \cite{yu2018cvpr}, which may make the model ``forget'' distant layers, and aggregating layers is of profound value to better fuse linguistic information at different levels of representation \cite{peters2018deep,shen2018dense,wang2018multilayer,wang2019learning,dou2018exploiting,dou2019dynamic}. Selectively aggregating different layer representations of the Transformer may further improve the performance.

In this paper, we propose to train Transformers with depth-wise LSTMs which regard outputs of stacked Transformer layers as steps in a time series and manage representation aggregation in and across layers. Our general motivation is that complex cross-layer information management offered by depth-wise LSTMs may bring about additional benefits over simple residual connections: LSTMs \cite{Hochreiter1997LSTM} have been shown to (i) avoid gradient explosion and vanishing, (ii) selectively learn what to remember and what to forget while ensuring convergence.

We explore the use of LSTMs to connect layers in stacked deep architectures for Transformers: we show how residual connections can be replaced by LSTMs connecting self-, cross- and masked self-attention layers. In contrast to standard LSTMs that process token sequences, we refer to the use of LSTMs in connecting stacked layers of deep architectures as ``depth-wise LSTMs''.

Our contributions are as follows:

\begin{itemize}
	\item We suggest that selectively aggregating different layer representations of the Transformer may improve the performance, and propose to use depth-wise LSTMs to connect stacked (sub-) layers of Transformers. We show how Transformer layer normalization and feed-forward sub-layers can be absorbed by depth-wise LSTMs, while connecting pure Transformer attention layers by depth-wise LSTMs (for Transformer encoder and decoder blocks), replacing residual connections.
	\item We show that the 6-layer Transformer using depth-wise LSTM can bring significant improvements in both WMT tasks and the challenging OPUS-100 multilingual NMT task. We show that depth-wise LSTM also has the ability to support deep Transformers with up to $24$ layers, and that the 12-layer Transformer using depth-wise LSTM already performs at the level of the 24-layer vanilla Transformer.
\end{itemize}

\section{Transformer with Depth-Wise LSTM}
\label{secdwla}

\subsection{Depth-Wise LSTM}

The computation of depth-wise LSTM is the same as the conventional LSTM except that depth-wise LSTM connects stacked Transformer layers instead of tokens in a token sequence as in conventional LSTMs. The gate mechanisms in the original LSTM are to enhance its ability in capturing long-distance relations and to address the gradient vanishing/exploding issue in sequence modeling. In our work, we regard the outputs of stacked layers as a ``vertical'' sequence, and utilize the same gate mechanisms to selectively aggregate information from stacked Transformer layer outputs and to address the gradient vanishing issue of deep Transformers. LSTMs are able to capture long-distance relationships: they can learn to selectively use the representations of distant tokens in the processing of a current input token in a sequence. In a sense, the layer-by-layer computations in Transformer encoder and decoder stacks are just such sequences where information from a Transformer layer $n-1$ is passed on to layer $n$. Our depth-wise LSTMs connect layers of multi-head attention information instead of token embeddings. Because of the different types of attention (self, cross and masked), we develop tailored ways of connecting (sub-) layers in encoder stacks and decoder stacks with depth-wise LSTMs.

\begin{figure}[t]
	\centering
	\includegraphics[width=0.8\columnwidth]{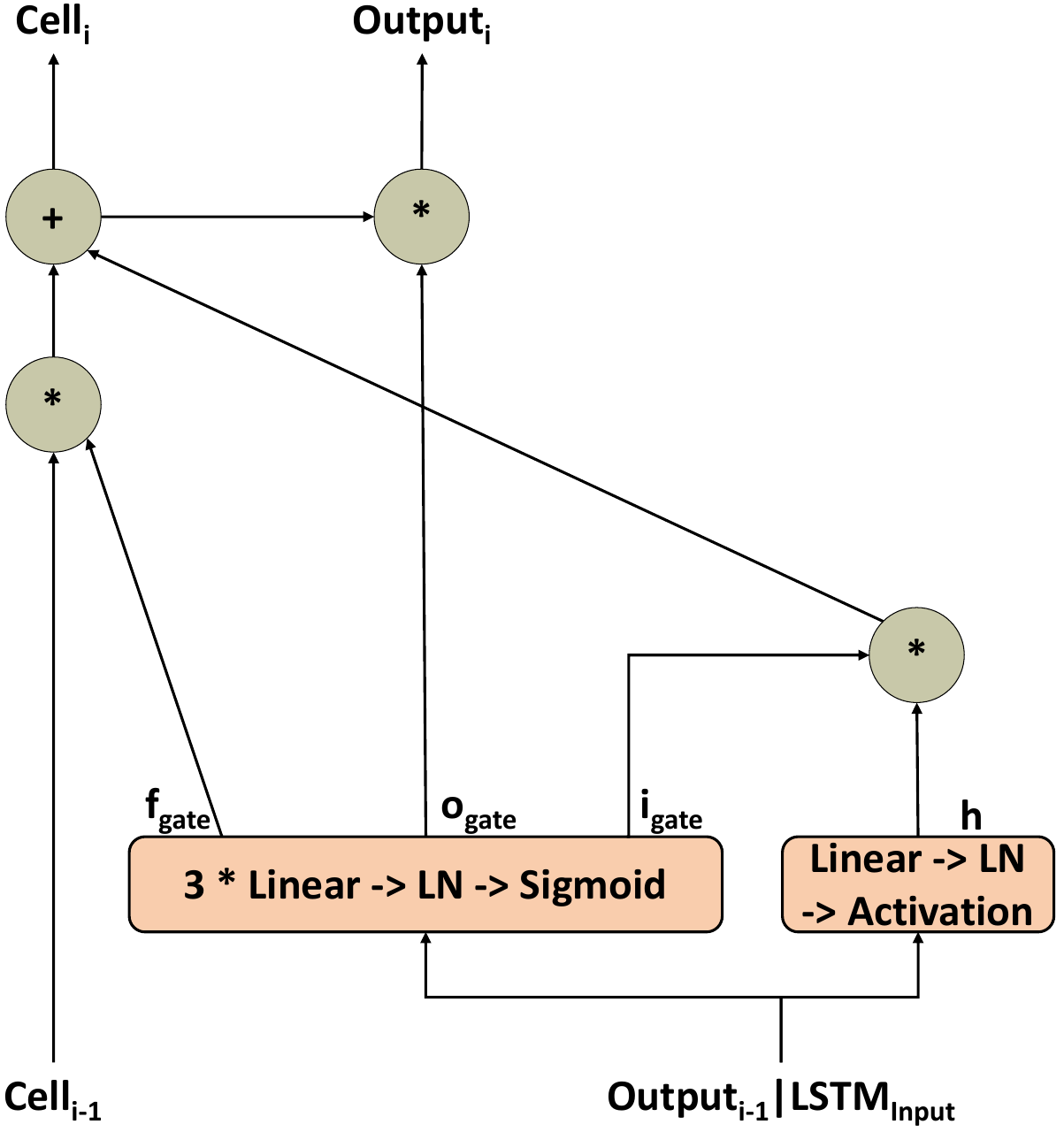}
	\caption{Depth-wise LSTM computation.}
	\label{fig:lstm}
\end{figure}

We equip our depth-wise LSTM with layer normalization. This has shown better performance as an LSTM-based NMT decoder \cite{chen2018best,xu2021multi} than vanilla LSTM. The computation graph of our depth-wise LSTM is shown in Figure~\ref{fig:lstm}.

The depth-wise LSTM concatenates the input from the current Transformer layer $LSTM_{Input}$ to the LSTM with the output of the LSTM from the previous layer $Output_{i-1}$:

\begin{equation}
{\rm{c}} = Output_{i - 1}|LSTM_{Input}
\end{equation}

\noindent where ``$|$'' indicates concatenation.

Next, the depth-wise LSTM computes three gates (input gate $i_{gate}$, forget gate $f_{gate}$ and output gate $o_{gate}$) and the hidden representation $h$ from the concatenated representation $c$:

\begin{equation}
i_{gate} = \sigma ({\mathop{\rm LN}\nolimits}({W_{i}}c + {b_{i}}))
\label{eqa:igate}
\end{equation}

\begin{equation}
f_{gate} = \sigma ({\mathop{\rm LN}\nolimits}({W_{f}}c + {b_{f}}))
\label{eqa:fgate}
\end{equation}

\begin{equation}
o_{gate} = \sigma ({\mathop{\rm LN}\nolimits}({W_{o}}c + {b_{o}}))
\label{eqa:ogate}
\end{equation}

\begin{equation}
h = {\mathop{\rm GeLU}\nolimits} ({\mathop{\rm LN}\nolimits}({W_{h}}c + {b_{h}}))
\label{eqa:lstmhid}
\end{equation}

\noindent where $W_*$ and $b_*$ are weight and bias parameters, $\sigma$ is the sigmoid activation function, ${\mathop{\rm LN}\nolimits}$ is the layer normalization.

We consider the role of the computation of the hidden state (Equation~\ref{eqa:lstmhid}) similar to the position-wise feed-forward sub-layer in each of the original Transformer encoder and decoder layers, and remove the feed-forward sub-layer from the original encoder and decoder layers when we connect them by our depth-wise LSTMs. The original Transformer uses a 2-layer feed-forward network. In an additional set of experiments we model these two layers in the hidden state of the depth-wise LSTM in terms of two weight matrices $W_{h1}$ and $W_{h2}$ but use the GLU activation function \cite{shazeer2020glu} for parameter efficiency, as shown in Equation~\ref{eqa:transformerhid} (compare Equation~\ref{eqa:lstmhid}):

\begin{equation}
h = W_{h2} {\mathop{\rm GLU}\nolimits} ({\mathop{\rm LN}\nolimits}({W_{h1}}c + {b_{h1}})) + b_{h2}
\label{eqa:transformerhid}
\end{equation}

After the computation of the hidden state, the cell state and the output of the LSTM unit are computed as:

\begin{equation}
Cell_i = Cell_{i - 1} * {f_{gate}} + h * {i_{gate}}
\label{eqa:lstmcell}
\end{equation}

\begin{equation}
Output_i = Cell_i * {o_{gate}}
\label{eqa:lstmout}
\end{equation}

\noindent where $*$ indicates element-wise multiplication.

As the depth-wise LSTM is computed across stacked Transformer layers and the token embeddings are already produced before computing the first encoder/decoder layer, we use the token embeddings as $Cell_0$ and $Output_0$.

The gate mechanisms (Equations~\ref{eqa:igate},~\ref{eqa:fgate},~\ref{eqa:ogate}) of the depth-wise LSTM can selectively learn to treat representations from different Transformer levels differently while guarding against vanishing and exploding gradients (Table~\ref{tab:ablrnn}).

We use depth-wise LSTM rather than a depth-wise multi-head attention network \cite{dou2018exploiting} with which we can build the NMT model solely based on the attention mechanism for two reasons: 1) we have to compute the stacking of Transformer layers sequentially as in sequential token-by-token decoding, and compared to the use of depth-wise LSTM of $O(n)$ complexity, depth-wise multi-head attention networks suffer from $O(n^2)$ complexity and they cannot be parallelized at the depth level. 2) the attention mechanism linearly combines representations with attention weights. Thus, it lacks the ability to provide the non-linearity compared to the LSTM, which we suggest is important.

\begin{figure}[t]
	\centering
	\includegraphics[width=0.5\columnwidth]{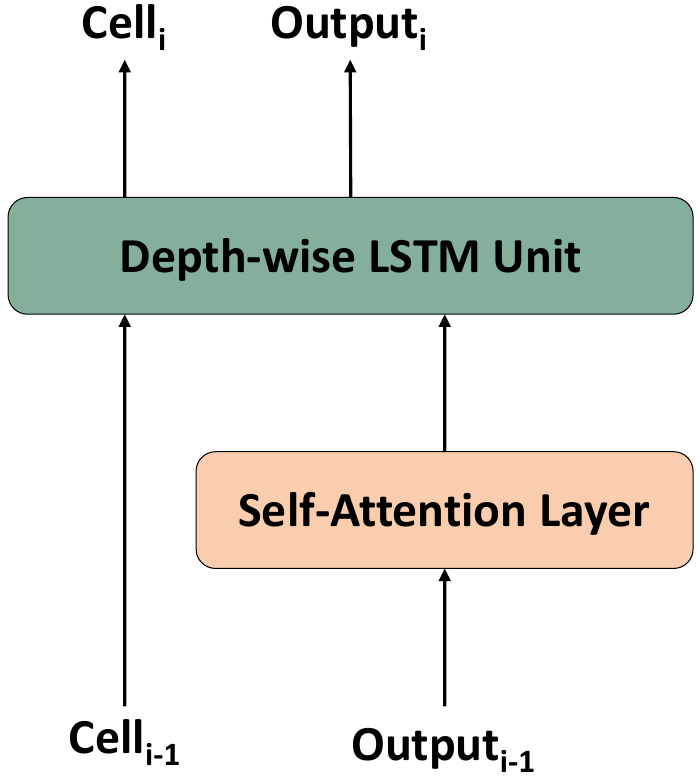}
	\caption{Encoder layer with depth-wise LSTM.}
	\label{fig:enc}
\end{figure}

\subsection{Encoder Layers Connected via Depth-Wise LSTMs}

Directly replacing residual connections with LSTM units will introduce a large amount of additional parameters and computation. Given that the task of computing the LSTM hidden state is similar to the feed-forward sub-layer in the original Transformer layers, we propose to replace the feed-forward sub-layer with the newly introduced LSTM unit, which only introduces one LSTM unit per layer, and the parameters of the LSTM can be shared across layers.

The original Transformer encoder layer only contains two sub-layers: the self-attention sub-layer based on the multi-head attention network and the 2-layer feed-forward network sub-layer.

The encoder layer with the depth-wise LSTM unit, as shown in Figure~\ref{fig:enc}, first performs the self-attention computation, then the depth-wise LSTM unit takes the self-attention results and the output and the cell state of the previous layer to compute the output and the cell state of the current layer.

\begin{figure}[t]
	\centering
	\includegraphics[width=0.8\columnwidth]{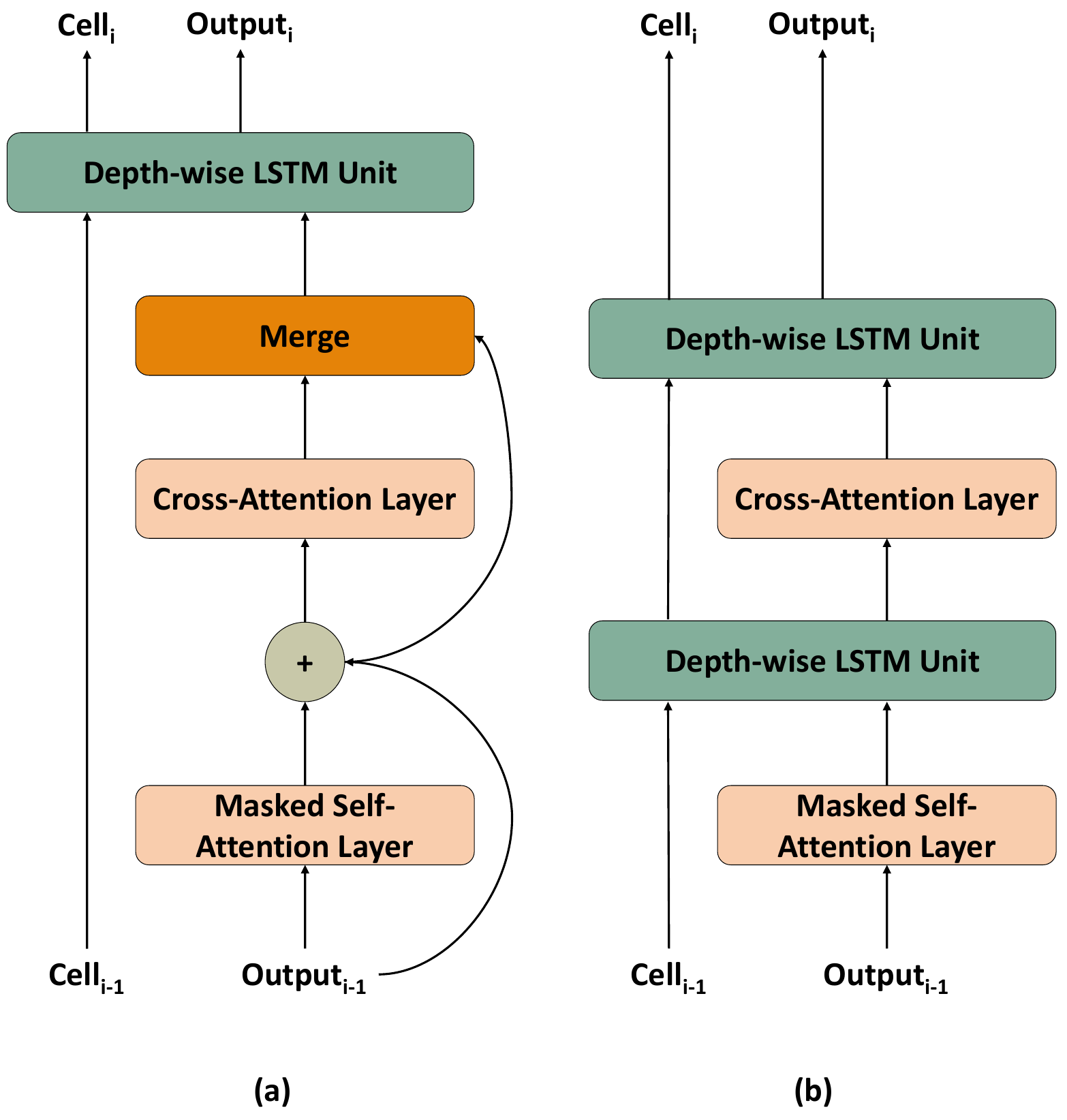}
	\caption{Decoder layer with depth-wise LSTM.}
	\label{fig:dec}
\end{figure}

\subsection{Decoder Layers Connected via Depth-Wise LSTMs}

Different from encoder layers, decoder layers involve two multi-head attention sub-layers: a masked self-attention sub-layer to attend the decoding history and a cross-attention sub-layer to attend information from the source side. Given that the depth-wise LSTM unit only takes one input, we introduce a merging layer to merge the outputs of these two sub-layers into one as the input to the LSTM unit. The architecture is shown in Figure~\ref{fig:dec} (a).

Specifically, the decoder layer with depth-wise LSTM first computes the masked self-attention sub-layer and the cross-attention sub-layer as in the original decoder layer, then it merges the outputs of these two sub-layers and feeds the merged representation into the depth-wise LSTM unit which also takes the cell and the output of the previous layer to compute the output of the current decoder layer and the cell state of the LSTM. We examine both element-wise addition and concatenation as merging operation.

Another way to take care of the outputs of these two sub-layers in the decoder layer is to replace their residual connections with two depth-wise LSTM sub-layers, as shown in Figure~\ref{fig:dec} (b). This leads to better performance (as shown in Table~\ref{tab:ablmerge}), but at the costs of more parameters and decoder depth in terms of sub-layers.

\section{Experiments}

We implemented our approach based on the Neutron implementation of the Transformer \cite{xu2019neutron}. To show the effects of depth-wise LSTMs on the 6-layer Transformer, we first conducted experiments on the WMT 14 English to German and English to French news translation tasks to compare with the Transformer baseline \citet{vaswani2017attention}. Additionally, we also examined the impact of our approach on deep Transformers and in a multilingual NMT task. The deep Transformer experiments were conducted on the WMT 14 English to German task and the WMT 15 Czech to English task following \citet{bapna2018training,xu2020lipschitz}, and the multilingual NMT experiments were performed on the challenging OPUS-100 dataset following \citet{zhang2020improving}. The concatenation of newstest 2012 and newstest 2013 was used for validation and newstest 2014 as test set for the WMT 14 English to German and English to French news translation tasks, and newstest 2013 as validation set for the WMT 15 Czech to English task. Newstest 2014 provided the test sets for both the WMT 14 English to German and the English to French task, and newstest 2015 was the test set for the Czech to English task.

\subsection{Settings}

We applied joint Byte-Pair Encoding \cite{sennrich2015neural} with $32k$ merging operations on all data sets to address the unknown word issue. We only kept sentences with a maximum of $256$ subword tokens for training. For fair comparison, we did not tune any hyperparameters but followed \citet{vaswani2017attention} for all experiment settings.

Though \citet{zhang2019improving,xu2020dynamically} suggest using a large batch size which may lead to improved performance, we only used a batch size of $25k$ target tokens (through gradient accumulation of small batches) to fairly compare with previous work \cite{vaswani2017attention,xu2020lipschitz}.

We used a beam size of $4$ for decoding, and evaluated tokenized case-sensitive BLEU with the averaged model of the last $5$ checkpoints for the Transformer Base setting and $20$ checkpoints for the Transformer Big setting saved at intervals of $1,500$ training steps. We also conducted significance tests \cite{koehn2004statistical}. To measure the efficiency of different settings, we sorted the WMT 14 En-De test set of $3003$ sentences by the number of tokens of the input sentence to reduce the number of padding tokens during batching, and tested the inference speed on a single RTX 4090 GPU (matrix multiplications were computed in FP16 precision for faster decoding), and reported the beam decoding speed (number of sentences per second).

\begin{table}[t]
	\centering
	\begin{tabular}{lll}
		\toprule
		Models & \multicolumn{1}{l}{En-De} & \multicolumn{1}{l}{En-Fr} \\
		\midrule
		Transformer Base  & 27.55  & 39.54 \\
		\multicolumn{1}{c}{with depth-wise LSTM}   & \textbf{28.53}$^\dag$  & \textbf{40.10}$^\dag$ \\
		\midrule
		Transformer Big   & 28.83 & 41.92 \\
		\multicolumn{1}{c}{with depth-wise LSTM}   & \textbf{29.58}$^\dag$ & \textbf{43.11}$^\dag$ \\
		\bottomrule
	\end{tabular}
	\caption{Results on WMT 14 En-De and En-Fr. $^\dag$ indicates $p<0.01$ in the significance test.}
	\label{tab:bleuall}
\end{table}

\subsection{Main Results}

We first examine the effects of our approach on the 6-layer Transformer on the WMT 14 English-German and English-French task to compare with \citet{vaswani2017attention}, and results are shown in Table~\ref{tab:bleuall}.

In our approach (``with depth-wise LSTM''), we used the 2-layer neural network for the computation of the LSTM hidden state (Equation~\ref{eqa:transformerhid}) and shared LSTM parameters across stacked encoder layers and different shared parameters across decoder layers for computing the LSTM gates (Equations~\ref{eqa:igate},~\ref{eqa:fgate},~\ref{eqa:ogate}). Details are provided in our ablation study.

Table~\ref{tab:bleuall} shows that our approach based on the depth-wise LSTM can obtain significant improvements on both tasks over the original Transformer with both the Transformer Base setting and the Transformer Big setting. In particular, significant improvements ($+1.19$ BLEU) obtained by our approach on the En-Fr task (trained on $\sim$$36$M sentence pairs) with the Transformer Big support the effectiveness of our approach in large-scale and challenging settings.

Our approach with the Transformer base setting brings about more improvements on the English-German task than that on the English-French task. We conjecture that maybe because the performance on the English-French task using a large dataset ($\sim$$36$M sentence pairs) may rely more on the capacity of the model (i.e. the number of parameters) than on the complexity of the modeling function (i.e. depth of the model, non-linearity strength per-layer, etc.). With the Transformer Big model which contains more parameters than the Transformer Base, the improvement on En-Fr ($+1.19$) is larger than that on En-De ($+0.75$), with $\sim$$4.5$M sentence pairs.

\begin{table}[t]
	\centering
	\begin{tabular}{lrrr}
		\toprule
		\multicolumn{1}{c}{Approaches} & \multicolumn{1}{c}{BLEU} & \multicolumn{1}{c}{Para.(M)} & \multicolumn{1}{c}{Speed} \\
		\midrule
		Transformer & 27.55 & 62.37 & 750.58 \\
		Depth-wise RNN & 23.24 & 68.67 & 737.60 \\
		Depth-wise LSTM & \textbf{28.53} & 70.25 & 674.96 \\
		\bottomrule
	\end{tabular}
	\caption{Ablation study of depth-wise approaches on WMT 14 En-De.}
	\label{tab:ablrnn}
\end{table}

\subsection{Ablation Study}
\label{subsec:expabl}

\begin{table*}[t]
	\centering
	\begin{tabular}{lrrrr}
		\toprule
		\multicolumn{1}{c}{LSTM FFN} & \multicolumn{1}{c}{Hidden size} & \multicolumn{1}{c}{BLEU} & \multicolumn{1}{c}{Para.(M)} & \multicolumn{1}{c}{Speed} \\
		\midrule
		1-layer (Eq.~\ref{eqa:lstmhid}) & 512 & 27.84 & 45.05 & 742.19 \\
        \midrule
		\multirow{2}[0]{*}{2-layer (Eq.~\ref{eqa:transformerhid})} & 2048 & \textbf{28.53} & 70.25 & 674.96 \\
		& 1586 & 28.20 & 62.37 & 683.67 \\
		\bottomrule
	\end{tabular}
	\caption{Ablation study of LSTM hidden computation on WMT 14 En-De.}
	\label{tab:ablffn}
\end{table*}

\begin{table}[t]
	\centering
	\small
	\begin{tabular}{lrrr}
		\toprule
		\multicolumn{1}{c}{Merging} & \multicolumn{1}{c}{BLEU} & \multicolumn{1}{c}{Para.(M)} & \multicolumn{1}{c}{Speed} \\
		\midrule
		Concat & 28.26 & 78.90 & 649.27 \\
		Add   & 28.53 & 70.25 & 674.96 \\
		\midrule
		$2$ Depth-wise LSTMs & \textbf{28.81} & 100.18 & 581.13 \\
		\bottomrule
	\end{tabular}
	\caption{Results of merging operations for decoder layer on WMT 14 En-De.}
	\label{tab:ablmerge}
\end{table}

\begin{table}[t]
	\centering
	\begin{tabular}{lrr}
		\toprule
		\multicolumn{1}{c}{Sharing} & \multicolumn{1}{c}{BLEU} & \multicolumn{1}{c}{Para.(M)} \\
		\midrule
		All   & 26.94 & 44.00 \\
		Gate  & \textbf{28.53} & 70.25 \\
		None  & 28.25 & 87.59 \\
		\bottomrule
	\end{tabular}
	\caption{Results of sharing LSTM parameters on WMT 14 En-De.}
	\label{tab:ablshare}
\end{table}

\begin{table*}[t]
	\centering
	\small
	\begin{tabular}{cccllrr}
		\toprule
		\multirow{2}[0]{*}{Models} & \multicolumn{2}{c}{Layers} & \multicolumn{1}{c}{\multirow{2}[0]{*}{En-De}} & \multicolumn{1}{c}{\multirow{2}[0]{*}{Cs-En}} & \multicolumn{1}{c}{\multirow{2}[0]{*}{Para.(M)}} & \multicolumn{1}{c}{\multirow{2}[0]{*}{Speed}} \\
		& \multicolumn{1}{l}{Encoder} & \multicolumn{1}{l}{Decoder} &       & & & \\
		\midrule
		\multicolumn{5}{l}{Transformer Base} \\
		\multicolumn{1}{l}{TA \cite{bapna2018training}$^*$} & \multicolumn{1}{c}{16} & \multicolumn{1}{c}{\multirow{3}[0]{*}{6}} & 28.39 & 29.36 & 93.87 & 711.78 \\
		\multicolumn{1}{l}{DLCL \cite{wang2019learning}} & \multicolumn{1}{c}{30} & & 29.3 & \multicolumn{1}{c}{\multirow{6}[0]{*}{None}} & 137.97 & 577.30 \\
  	\multicolumn{1}{l}{ODE \cite{li2022ode}} & \multicolumn{1}{c}{24} & & 30.29 & & 119.17 & 565.86 \\
		\multicolumn{1}{l}{Layer Aggregation \cite{dou2018exploiting}} & \multicolumn{2}{c}{6} &  28.63 & & 111.10 & 667.57 \\
		\multicolumn{1}{l}{EM Routing \cite{dou2019dynamic}} & \multicolumn{2}{c}{6} &  28.81 & & 144.80 & 561.28 \\
		\multicolumn{1}{l}{SDU \cite{chai2020highway}$^*$} & \multicolumn{2}{c}{6} & 28.22 & & 78.13 & 664.20 \\
		\multicolumn{1}{l}{Luna \cite{ma2021luna}} & \multicolumn{2}{c}{6} & 27.8 & & 77.60 & None \\
		\multicolumn{1}{l}{DSI \cite{zhang2019improving}} & \multicolumn{2}{c}{20} & 28.67 & & 149.54 & 298.50 \\
		\multicolumn{1}{l}{LCPI \cite{xu2020lipschitz}} & \multicolumn{2}{c}{24} & 29.20 & 29.88 & 194.66 & 229.90 \\
		\midrule
		\multicolumn{5}{l}{Transformer Big} \\
		\multicolumn{1}{l}{Layer Aggregation \cite{dou2018exploiting}} & \multicolumn{2}{c}{6} &  29.21 & \multirow{4}[0]{*}{None} & 356.38 & 264.55 \\
		\multicolumn{1}{l}{EM Routing \cite{dou2019dynamic}} & \multicolumn{2}{c}{6} &  28.97 &  & 490.38 & 221.70 \\
		\multicolumn{1}{l}{MC \cite{wei2020multiscale}} & \multicolumn{2}{c}{18} & 30.56 &  & 798.23 & 70.37 \\
  	\multicolumn{1}{l}{ODE \cite{li2022ode}} & \multicolumn{1}{c}{12} & \multicolumn{1}{c}{6} & 30.77 & & 288.46 & 315.91 \\
		\midrule
		\multirow{5}[0]{*}{Transformer Base} & \multicolumn{2}{c}{\ 3} & 26.36 & 27.91 & 40.33 & 1209.62 \\
		& \multicolumn{2}{c}{\ 6} & 27.55  & 28.40 & 62.37 & 750.58 \\
		& \multicolumn{2}{c}{12} & 28.12  & 29.38 & 106.47 & 429.00 \\
		& \multicolumn{2}{c}{18} & 28.60  & 29.61 & 150.57 & 299.81 \\
		& \multicolumn{2}{c}{24} & 29.02  & 29.73 & 194.66 & 229.90 \\
		\midrule
		\multirow{5}[0]{*}{Transformer Base with depth-wise LSTM} & \multicolumn{2}{c}{\ 3} & 27.38 & 28.26 & 46.63 & 1121.16 \\
		& \multicolumn{2}{c}{\ 6} & 28.53 & 29.15 & 70.25 & 674.96 \\
		& \multicolumn{2}{c}{12} & 29.26  & 29.64 & 122.23 & 379.83 \\
		& \multicolumn{2}{c}{18} & \textbf{29.41}  & \textbf{30.27} & 172.63 & 277.21 \\
		& \multicolumn{2}{c}{24} & 29.18  & 30.02 & 223.02 & 202.40 \\
		\midrule
		\multicolumn{1}{l}{Transformer Big with depth-wise LSTM} & \multicolumn{2}{c}{12} & \textbf{30.69} & \textbf{30.57} & 452.04 & 181.58 \\
		\multicolumn{1}{l}{\ \ \ \ + experiment settings of \citet{li2022ode}} & \multicolumn{1}{c}{\multirow{2}[0]{*}{12}} & \multicolumn{1}{c}{\multirow{2}[0]{*}{6}} & \textbf{31.12} & \textbf{31.25} & 338.75 & 316.15 \\
		\multicolumn{1}{l}{\ \ \ \ \ \ \ \ + 1-layer LSTM FFN (Eq.~\ref{eqa:lstmhid})} & & & 30.83 & 30.96 & 288.41 & 363.60 \\
		\bottomrule
	\end{tabular}
	\caption{Results of Deep Transformers. ``*'' indicates reproduction of the approach.}
	\label{tab:deep}
\end{table*}

We conducted ablation studies on the WMT 14 En-De task with the Base setting.

Considering that the layer stacks of the 6-layer Transformer are not that deep and vanilla RNNs can play a similar role as LSTMs, is it possible to train the model with a depth-wise RNN rather than the depth-wise LSTM? We first study using different approaches (Transformer, the depth-wise RNN and the depth-wise LSTM) for the 6-layer Transformer, and results are shown in Table~\ref{tab:ablrnn}.

When using the depth-wise RNN, the architecture is quite similar to the standard Transformer layer without residual connections but using the concatenation of the input to the encoder/decoder layer with the output(s) of attention layer(s) as the input to the last FFN sub-layer. Table~\ref{tab:ablrnn} shows that the 6-layer Transformer with the depth-wise RNN is able to converge, but its performance is much worse than the model with the depth-wise LSTM (and also much worse than the vanilla Transformer) with depth-wise LSTM outperforming the vanilla Transformer, suggesting the importance of the gating mechanisms of the depth-wise LSTM. The decoding speed of our baseline vanilla Transformer implementation ($750.58$ sentences/s) is quite fast, and is $1.12$ times as fast as the depth-wise LSTM approach, but our approach leads to a higher BLEU score than the baseline, and as shown in Table~\ref{tab:deep}, our approach indeed requires fewer parameters and brings about faster decoding speed than the vanilla Transformer for a comparable BLEU score.

Next, we study the effects of two types of computations for the LSTM hidden state in Equations~\ref{eqa:lstmhid} and~\ref{eqa:transformerhid} on the performance on the WMT 14 En-De task. Results are shown in Table~\ref{tab:ablffn}.

Table~\ref{tab:ablffn} shows that a 2-layer feed-forward neural network (Equation~\ref{eqa:transformerhid}) in the depth-wise LSTM outperforms the original computation of the LSTM hidden state which uses only one layer (Equation~\ref{eqa:lstmhid}), which is consistent with intuition. However, even with only one layer for the hidden state computation and with $27.77\%$ fewer parameters ($45.05$M against $62.37$M), depth-wise LSTM (Equation~\ref{eqa:lstmhid}) still slightly outperforms the vanilla Transformer baseline in BLEU ($27.84$ against $27.55$), suggesting that the improvements from using depth-wise LSTMs are not just due to the increased amount of parameters. The 1-layer LSTM FFN model also archieves a comparable decoding speed compared to the baseline ($742.19$ v.s. $750.58$). When we reduce the hidden dimension of Equation~\ref{eqa:transformerhid} to $1586$, which results in approximately the same number of parameters as the standard Transformer, depth-wise LSTM still outperforms the baseline by $+0.65$ BLEU.

We also study the merging operations, concatenation, element-wise addition, and the use of 2 depth-wise LSTM sub-layers, to combine the masked self-attention sub-layer output and the cross-attention sub-layer output in decoder layers. Results are shown in Table~\ref{tab:ablmerge}.

Table~\ref{tab:ablmerge} shows that, even though this is counter-intuitive, element-wise addition (with fewer parameters) empirically results in slightly higher BLEU than the concatenation operation. Furthermore, even though using 2 depth-wise LSTM sub-layers connecting cross- and masked self-attention sub-layers leads to the highest BLEU score, showing the advantage of fully replacing residual connections with depth-wise LSTMs, it also introduces more parameters and increases the decoder depth in terms of sub-layers. For fair comparison, we use the simpler element-wise addition operation in our experiments by default.

As the number of Transformer layers is pre-specified, the parameters of the depth-wise LSTM can either be shared across layers or be independent. Table~\ref{tab:ablffn} documents the importance of the capacity of the module for the hidden state computation, and sharing the module is likely to hurt its capacity. We additionally study to share only parameters for gate computation (Equations~\ref{eqa:igate},~\ref{eqa:fgate},~\ref{eqa:ogate}) and to share all parameters (i.e. parameters for both the computation of gates and of the hidden state). Results are shown in Table~\ref{tab:ablshare}.

Table~\ref{tab:ablshare} shows that: 1) Sharing parameters for the computation (Equation~\ref{eqa:transformerhid}) of the depth-wise LSTM hidden state significantly hampers performance, which is consistent with our conjecture. 2) Sharing parameters for the computation of gates (Equations~\ref{eqa:igate},~\ref{eqa:fgate},~\ref{eqa:ogate}) leads to slightly higher BLEU with fewer parameters introduced than without sharing them (``None'' in Table~\ref{tab:ablshare}). Thus, in the other experiments, we bind parameters for the computation of LSTM gates across stacked layers by default.

\subsection{Deep Transformers}

We examine whether depth-wise LSTM has the ability to ensure the convergence of deep Transformers and measure performance on the WMT 14 English to German task and the WMT 15 Czech to English task following \citet{bapna2018training,xu2020lipschitz}, and compare our approach with the pre-norm Transformer in which residual connections are not normalized by layer normalization. To compare with the previous studies, we replace the English to French task with the Czech to English task with $\sim$$15$M sentence pairs. The $4.5$M dataset of the En-De task is not small, and the Cs-En data that has more than $15$M sentence pairs is even larger, and can be considered a large-scale dataset. Together with the English-French experiment (Table~\ref{tab:bleuall}), this allows us to assess the effectiveness of our approach with large datasets and deep Transformers. For fairness and reliable comparisons across all our experiments, we strictly followed the experiment settings of \citet{vaswani2017attention} by default, without using relative positional encoding \cite{shaw2018self}, dense connections, larger number of warm up steps, and larger batch sizes, although several previous studies \cite{wang2019learning,zhang2019improving,li2020shallow,li2022ode} employ some or all of these different settings for higher BLEU scores. Results are shown in Table~\ref{tab:deep}.

Table~\ref{tab:deep} shows that though the BLEU improvements start saturating with deep depth-wise LSTM Transformers of more than $12$ layers, depth-wise LSTM is able to ensure convergence of up to $24$ layer Transformers. The experiments also show that the size differences between these datasets did not lead to differences in optimization.

Notably, on the En-De task, the 12-layer Transformer with depth-wise LSTM already outperforms the 24-layer vanilla Transformer, suggesting efficient use of layer parameters. On the Cs-En task, the 12-layer model with depth-wise LSTM performs on a par with the 24-layer baseline. Unlike in the En-De task, increasing depth over the 12-layer Transformer can still achieve some BLEU improvements, with the 18-layer model resulting in the best performance. We conjecture that this is probably because the data set of the Cs-En task ($\sim$$15$M) is larger than that of the En-De task ($\sim$$4.5$M), and increasing the depth of the model for the Cs-En task also increases its number of parameters and capacity. For the En-De task, the 12-layer Transformer with depth-wise LSTM may already provide both sufficient complexity and capacity for the data set.

It is a common problem that increasing the depth does not always lead to better performance, whether with residual connections \cite{li2022works} or other previous studies on deep Transformers \cite{bapna2018training,wang2019learning,li2022ode}, and the use of wider models is the usual method of choice for further improvements. Although for the Base Transformer model our approach does not lead to significant improvements for models deeper than $18$ layers, we argue that the 18-layer Transformer Base is not the performance limit of our approach, because we may increase the width of the model in addition to the depth. As shown in Table~\ref{tab:deep}, the 12-layer Transformer Big with depth-wise LSTM is able to achieve further improvements over Transformer Base models, and using fewer layers and parameters achieves performance on par with \citet{wei2020multiscale}. Using relative positional encoding, larger batch sizes, etc., following the experiment settings of \citet{li2022ode} can also lead to better performance with our approach.

As for the costs, the decoder depth has a strong impact on inference speed, as the decoder has to be computed once for each decoding step during auto-regressive decoding \cite{kasai2020deep,xu2021probing}, and the use of only deep encoders \cite{bapna2018training,wang2019learning,li2022ode,chai2020highway} normally leads to faster inference speed than using both a deep encoder and a deep decoder. But in general, Table~\ref{tab:deep} shows that our approach uses fewer parameters and leads to faster decoding speed than the baselines to obtain a comparable BLEU score, showing the efficiency of our method.

\begin{table*}[t]
  \centering
    \begin{tabular}{ccrrr}
    \toprule
    \multicolumn{1}{c}{Models} & Direction & \multicolumn{1}{c}{BLEU$_{94}$} & \multicolumn{1}{c}{WR} & \multicolumn{1}{c}{BLEU$_{4}$} \\
    \midrule
	\multirow{2}[0]{*}{Transformer} & En$\rightarrow$xx & 18.75 & \multirow{2}[0]{*}{-} & 14.73 \\
          & xx$\rightarrow$En & 27.02 &       &  22.50 \\
    \midrule
    \multirow{2}[0]{*}{Transformer + LALN + LALT \cite{zhang2020improving}} & En$\rightarrow$xx & 20.81 & \multirow{2}[0]{*}{-} & 17.45 \\
          & xx$\rightarrow$En & 27.22 &       & 23.30 \\
    \midrule
    \multirow{2}[0]{*}{Depth-wise LSTM} & En$\rightarrow$xx  & \textbf{23.38} & 98.94 & \textbf{20.47} \\
          & xx$\rightarrow$En & \textbf{28.41} & 79.79 & \textbf{26.68} \\
    \bottomrule
    \end{tabular}
  \caption{Results of multilingual NMT.}
  \label{tab:opus}
\end{table*}

\subsection{Multilingual NMT}

Multilingual translation uses a single model to translate between multiple language pairs \cite{firat2016multi,johnson2017google,aharoni2019massively}. Model capacity has been found crucial for massively multilingual NMT to support language pairs with varying typological characteristics \cite{zhang2020improving,xu2021modeling}. Using model layers efficiently with depth-wise LSTMs is likely to benefit multilingual NMT.

To test the effectiveness of depth-wise LSTMs in the multilingual setting, we conducted experiments on the challenging massively many-to-many translation task on the OPUS-100 corpus \cite{tiedemann2012parallel,aharoni2019massively,zhang2020improving}. We tested the performance of 6-layer models following the experiment settings of \citet{zhang2020improving} for fair comparison. We adopted BLEU \cite{papineni2002bleu} for translation evaluation with the SacreBLEU toolkit \cite{post2018call}. \footnote{BLEU+case.mixed+numrefs.1+smooth.exp+tok.13a\\+version.1.4.1} We report average BLEU over $94$ language pairs BLEU$_{94}$, win ratio WR (\%) compared to \citet{zhang2020improving}, average BLEU over $4$ selected typologically different target languages with varied training data sizes (de, zh, br, te) BLEU$_4$. Results are shown in Table~\ref{tab:opus}.

Compared to the baseline \cite{zhang2020improving}, Table~\ref{tab:opus} shows that: 1) our approach can lead to $+3.02$ and $+3.38$ BLEU improvements on average in the En$\rightarrow$xx and xx$\rightarrow$En directions respectively in the evaluation over 4 typologically different languages, and 2) using depth-wise LSTM is able to bring about $+2.57$ and $+1.19$ BLEU improvements on average when translating English to 94 languages and translating them into English respectively. Given that the one-to-many translation task requires more model capacity than the many-to-one translation task \cite{naveen2019massively}, the larger average BLEU improvements and a higher win ratio of $98.94\%$ (93 of 94 languages) in the En$\rightarrow$xx direction than in the xx$\rightarrow$En direction demonstrate the effectiveness of our approach especially when model capacity is crucial, suggesting the more effective use of model parameters with depth-wise LSTMs than vanilla Transformer.

\subsection{Efficiency Discussion}

Despite the depth-wise LSTM Transformers having more non-linear operations than the standard Transformer, we suggest that it is more efficient.

In our deep Transformer experiments, Table~\ref{tab:deep} shows that our depth-wise LSTM Transformer with fewer layers, parameters and computations can lead to competitive/better performance and faster decoding speed than vanilla Transformers with more layers but a similar BLEU score, and the depth-wise LSTM Transformer is in fact more efficient as we need fewer layers to achieve comparable performance.

In the multilingual NMT task which relies heavily on the model capacity, Table~\ref{tab:opus} shows that the use of depth-wise LSTM can bring about $+2.52$ BLEU improvements on average when translating English to 94 languages.

In Table~\ref{tab:ablffn}, we reduce the 2-layer FFN of the Transformer with depth-wise LSTM to only one layer with significantly fewer hidden units ($2048 \rightarrow 512$), this saves a large number of parameters and computations, and our approach with $45.05$M parameters still slightly outperforms the baseline residual Transformer with $62.37$M parameters (Table~\ref{tab:ablrnn}).

Our depth-wise LSTM Transformer Base performs on a par with the previous layer aggregation work \cite{dou2018exploiting} on the WMT 14 En-De task. However, our model only contains \textbf{$70.25$}M parameters while \citet{dou2018exploiting} involves \textbf{$111$}M parameters.

\section{Related Work}

\citet{he2016deep} present the residual learning framework to ease the training of deep neural networks. \citet{srivastava2015highway} propose the highway network which contains a transform gate and a carry gate to control the produced output and the input. \citet{chai2020highway} propose a highway Transformer with a self-gating mechanism for language models. However, our work is significantly different from theirs in two aspects. First, residual connections are still kept in their model. Second, their architecture does not use any mechanisms to track long-distance dependencies between stacked layers compared to depth-wise LSTM in our work.

\paragraph{Layer Aggregation} \citet{yu2018cvpr} suggest that skip connections are ``shallow'' themselves, and only fuse by simple, one-step operations, and therefore \citet{yu2018cvpr} augment standard architectures with deeper aggregation to better fuse information across layers to improve recognition and resolution. \citet{shen2018dense} propose a densely connected NMT architecture to create new features with dense connections. \citet{wang2018multilayer} propose a multi-layer representation fusion approach to learning a better representation from the layer stack. \citet{dou2018exploiting} simultaneously expose all layer representations with layer aggregation. \citet{dou2019dynamic} propose to use routing-by-agreement strategies to aggregate layers dynamically.

\paragraph{Deep NMT} \citet{zhou2016deep} introduce fast-forward connections and an interleaved bi-directional architecture for stacking LSTM layers. \citet{wang2017deep} propose a Linear Associative Unit to reduce the gradient propagation path inside the recurrent unit.

\paragraph{Deep Transformers} For the convergence of deep Transformers, \citet{bapna2018training} propose the Transparent Attention mechanism which allows each decoder layer to attend weighted combinations of all encoder layer outputs. \citet{wang2019learning} present the Dynamic Linear Combination of Layers approach that additionally aggregates shallow layers' outputs for each encoder layer. \citet{wu2019depth} propose a two-stage approach. \citet{wei2020multiscale} introduce a depth-wise GRU to additionally aggregate outputs of all encoder layers for the top decoder layer, but residual connections are still kept. \citet{zhang2019improving} and \citet{xu2020lipschitz} propose the layer-wise Depth-Scaled Initialization approach and the Lipschitz constrained parameter initialization approach, respectively, to reduce the standard deviation of layer normalization inputs and to ensure the functionality of residual connection. \citet{kasai2020deep,xu2021probing} propose to accelerate decoding by using deep encoders and shallower decoders. \citet{li2022ode} design an ODE Transformer which is analogous to the Runge-Kutta method. \citet{hao2022optimizing} present approaches to exploring hyperparameters of deep Transformers for low-resource NMT with shallow Transformers.

Regarding parameter efficiency for NMT, \citet{wu2018pay} present lightweight and dynamic convolutions. \citet{ma2021luna} approximate softmax attention with two nested linear attention functions. These methods are orthogonal to our work and it should be possible to combine them with our approach.

\section{Conclusion}

In this paper, we replace residual connections of the Transformer with depth-wise LSTMs, to selectively manage the representation aggregation of layers benefiting performance while ensuring convergence of the Transformer. Specifically, we show how to integrate the computation of multi-head attention networks and feed-forward networks with the depth-wise LSTM for the Transformer.

Our experiments with the 6-layer Transformer show that our approach using depth-wise LSTM can achieve significant BLEU improvements in both WMT news translation tasks and the very challenging OPUS-100 many-to-many multilingual translation task over baselines. Our deep Transformer experiments demonstrate that: 1) the depth-wise LSTM approach ensures that deep Transformers with up to $24$ layers converge, 2) the 12-layer Transformer using depth-wise LSTM already performs on a par with the 24-layer vanilla Transformer, suggesting more efficient usage of per-layer parameters with our depth-wise LSTM approach than the baseline.

\section{Acknowledgements}

We thank anonymous reviewers for their insightful comments. Hongfei Xu and Yang Song acknowledge the support of the National Natural Science Foundation of China (Grant No. 62306284), China Postdoctoral Science Foundation (Grant No. 2023M743189), and the Natural Science Foundation of Henan Province (Grant No. 232300421386). Josef van Genabith and Hongfei Xu are supported by the German Federal Ministry of Education and Research (BMBF) under funding code 01IW20010 (CORA4NLP). Deyi Xiong is partially supported by the Key Research and Development Program of Yunnan Province (No. 202203AA080004).

\nocite{*}
\section{Bibliographical References}\label{sec:reference}

\bibliographystyle{lrec-coling2024-natbib}
\bibliography{lrec-coling2024-example}

\end{document}